\def\year{2020}\relax
\documentclass[letterpaper]{article} 
\usepackage{aaai20}  
\usepackage{times}  
\usepackage{helvet} 
\usepackage{courier}  
\usepackage[hyphens]{url}  
\usepackage{graphicx} 
\usepackage{graphicx}  
\usepackage{amsmath}
\usepackage{multirow}
\usepackage{multicol}
\usepackage{array, boldline, makecell, booktabs}
\usepackage{amssymb}
\newcommand{\equationref}[1]{Eq.~\ref{#1}}
\newcommand{\tabref}[1]{Tab.~\ref{#1}}
\newcommand{\figref}[1]{Fig.~\ref{#1}}
\newcommand{\eg}[1]{\textit{e.g.}}
\newcommand{\ie}[1]{\textit{i.e.}}

\DeclareMathOperator*{\Modot}{\text{\raisebox{0.25ex}{\scalebox{0.6}{$\bigodot$}}}}

\title{Motion-Attentive Transition for Zero-Shot Video Object Segmentation}
\author{Tianfei Zhou\textsuperscript{\rm 1,2}, Shunzhou Wang\textsuperscript{\rm 2}, Yi Zhou\textsuperscript{\rm 1}, Yazhou Yao\textsuperscript{\rm 3}, Jianwu Li\textsuperscript{\rm 2}\thanks{Corresponding author: Jianwu Li(ljw@bit.edu.cn)}, Ling Shao\textsuperscript{\rm 1} \\
	\textsuperscript{\rm 1}Inception Institute of Artifical Intelligence, UAE \\ \textsuperscript{\rm 2}Beijing Key Laboratory of Intelligent Information Technology, \\ School of Computer Science and Technology, Beijing Institute of Technology, China \\ \textsuperscript{\rm 3}School of Computer Science and Engineering, Nanjing University of Science and Technology, China \\
\url{https://github.com/tfzhou/MATNet}}

\begin{document}
	
	\maketitle
	
	\begin{abstract}
In this paper, we present a novel Motion-Attentive Transition Network (MATNet) for zero-shot video object segmentation, which provides a new way of leveraging motion information to reinforce spatio-temporal object representation. An asymmetric attention block, called Motion-Attentive Transition (MAT), is designed within a two-stream encoder, which transforms appearance features into motion-attentive representations at each convolutional stage. In this way, the encoder becomes deeply interleaved, allowing for closely hierarchical interactions between object motion and appearance. This is superior to the typical two-stream architecture, which treats motion and appearance separately in each stream and often suffers from overfitting to appearance information. Additionally, a bridge network is proposed to obtain a compact, discriminative and scale-sensitive representation for multi-level encoder features, which is further fed into a decoder to achieve segmentation results. Extensive experiments on three challenging public benchmarks (\ie, DAVIS-16, FBMS and Youtube-Objects) show that our model achieves compelling performance against the state-of-the-arts.
\end{abstract}

\begin{figure*}
	\centering
	\includegraphics[width=\linewidth]{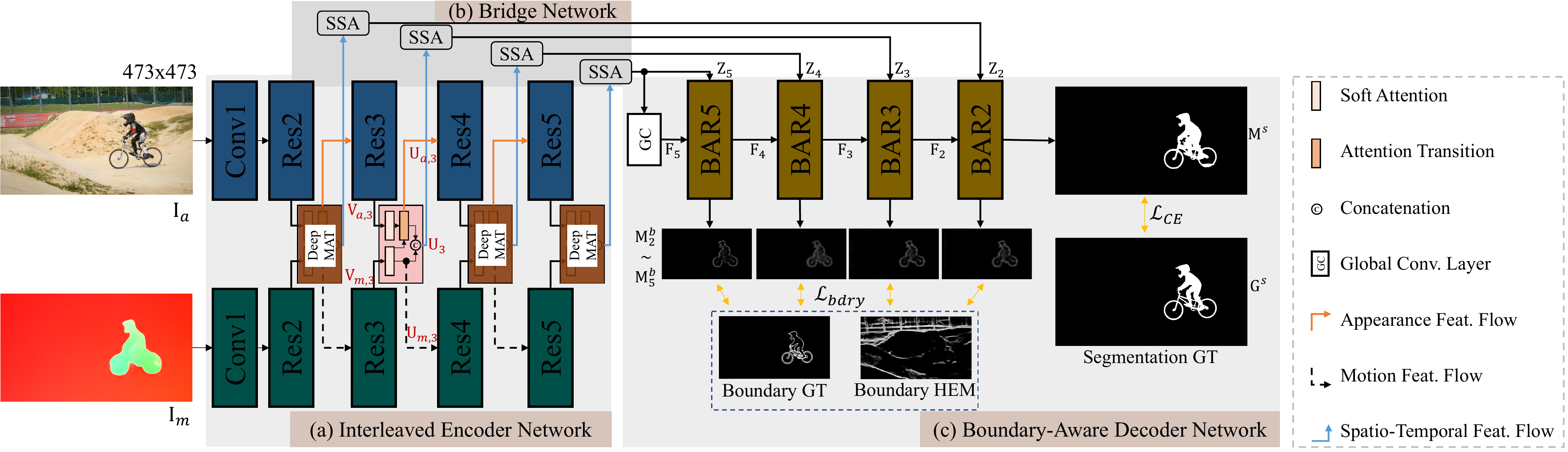}
	\caption{Pipleline of MATNet.
		The frame $\mathbf{I}_a$ and flow $\mathbf{I}_m$ are first input into the interleaved encoder to extract multi-scale spatio-temporal features $\mathbf{U}_i$.
		At each residual stage, we break the original information flow in ResNet.
		Instead, a deep MAT block is proposed to create a new interleaved information flow by simultaneously considering motion $\mathbf{V}_{m,i}$ and appearance $\mathbf{V}_{a,i}$.
		$\mathbf{U}_i$ is further fed into the decoder via the bridge network to obtain boundary results $\mathbf{M}_2^b \sim \mathbf{M}_5^b$ and the segmentation $\mathbf{M}^s$.}
	\label{fig:flow}
\end{figure*}

\section{Introduction}

The task of automatically segmenting primary object(s) from videos has gained significant attention in recent years, and has a powerful impact in many areas of computer vision, including surveillance, robotics and autonomous driving. However, due to the lack of human intervention, in addition to the common challenges posed by video data (\eg, appearance variations, scale changes, background clutter),  the task faces great difficulties in accurately discovering the most distinct objects throughout a video sequence.
Early non-learning methods typically address this using handcrafted features, \eg, motion boundary~\cite{papazoglou2013fast}, saliency~\cite{wang2015saliency} and point trajectories~\cite{ochs2013segmentation}. More recently,  research has turned towards the deep learning paradigm, with several studies attempting to fit this problem into a \textit{zero-shot} solution~\cite{ventura2019rvos,wang2019zero}. These methods generally learn a powerful object representation from large-scale training data and then adapt the models to test videos \textit{without any annotations}.

Even before the era of deep learning, \textit{object motion} has always been considered as an informative cue for automatic video object segmentation.  This is largely inspired by the remarkable capability of motion perception in the human visual system (HVS)~\cite{treisman1980feature,mital2013low}, which can quickly orient attentions towards moving objects in dynamic scenarios.  In fact, human beings are more sensitive to moving objects than static ones, even if the static objects are strongly contrasted against their surroundings.  Nevertheless, motion does not work alone. Recent studies~\cite{CLOUTMAN2013251} have revealed that, in HVS, the dorsal pathway (for motion perception) has a faster activation response than the ventral pathway (for objectness/semantic perception), and tends to send signals to the ventral pathway along multi-level connections to prompt it to focus on processing the most salient objects. This multi-modal system enables human beings to focus on the moving parts of objects first and then transfer the attention to appearance for the whole picture.
%
Thus, it is desirable to take this biological mechanism into account and incorporate object motion into appearance learning, in a hierarchical way, for more effective spatio-temporal object representation.

By considering information flow from motion to appearance,  we can alleviate ambiguity in object appearance (\eg, visually similar to the surroundings),  thus easing the pressure in representation learning of objects. However, in the context of deep learning, most segmentation models do not leverage this potential. Many approaches~\cite{tokmakov2017learningmotion,perazzi2017learning,jain2017fusionseg,cheng2017segflow} simply treat motion cues as equal to appearance cues and
learn to directly map optical flow to the corresponding segmentation mask. A few methods~\cite{xiao2018monet,li2018unsupervised} have attempted to enhance object representation with motion; however, they rely on complex heuristics and only operate at a single scale, ignoring the critical hierarchical structure.

Motivated by these observations, we propose a Motion-Attentive Transition Network (MATNet) for zero-shot video object segmentation (ZVOS) within an encoder-bridge-decoder framework, as shown in~\figref{fig:flow}. The core of MATNet is a deeply interleaved two-stream encoder which not only inherits the superiorities of two-stream models for multi-modal feature learning, but also progressively transfers intermediate motion-attentive features to facilitate appearance learning. The transition is carried out by multiple Motion-Attentive Transition (MAT) blocks. Each block takes as input the intermediate features of both the input image and optical flow map at a convolutional stage. Inside the block, we build an asymmetric attention mechanism that first infers regions of interest based on optical flow, and then transfers the inference to provide better selectivity for appearance features. Each block outputs the attentive appearance and motion features for the following convolutional stage.

%
In addition, our decoder accepts learnt features from the encoder as inputs and progressively refines coarse features scale-by-scale to obtain accurate segmentation. The decoder consists of multiple Boundary-Aware Refinement (BAR) blocks organized in a cascaded manner.  Each BAR explicitly exploits  multi-scale features and conducts segmentation inference with the assistance of object boundary prediction to obtain results with a finer structure.
%
%
Moreover, instead of directly connecting the encoder and decoder via skip connections,  we present the Scale-Sensitive Attention (SSA) to adaptively select and transform encoder features. Specifically, SSA, which is added to each pair of encoder and decoder layers,  consists of a two-level attention scheme in which the local-level attention serves to select a focused region, while the global-level one helps to re-calibrate features for objects at different scales.

%



MATNet can be easily instantiated with various backbones, and optimized in an end-to-end manner. We evaluate it on three popular video object segmentation benchmarks, \ie, DAVIS-16~\cite{perazzi2016benchmark}, FBMS~\cite{ochs2013segmentation}, and Youtube-Objects~\cite{prest2012learning}, and claim state-of-the-art performance.

\section{Related Work}

\textbf{Automatic Video Object Segmentation.}
Automatic, or unsupervised, video object segmentation aims to segment conspicuous and eye-catching objects without any human intervention. Traditional methods require no training data and typically design heuristic assumptions (\eg, motion boundary~\cite{papazoglou2013fast}, objectness~\cite{faktor2014video,zhou2016video,li2017learning}, saliency~\cite{wang2015saliency} and long-term point trajectories~\cite{ochs2013segmentation,keuper2015motion,ochs2011object}) for segmentation. In recent years, benefitting from the establishment of large datasets~\cite{perazzi2016benchmark,xu2018youtube},  many approaches~\cite{jain2017fusionseg,tokmakov2017learningmotion,li2018unsupervised,fan2018salient,lu2019see,hu2018unsupervised,Wang_2019_CVPR,ventura2019rvos,tokmakov2019learning,li2018instance,li2018unsupervised,song2018pyramid,faisal2019exploiting,fan2019shifting} propose to solve this task with zero-shot solutions and improve the performance greatly.
%

Among them, a large number of approaches utilize motion because of its complementary role to object appearance. They typically adopt heuristic methods to fuse motion and appearance cues~\cite{tokmakov2017learningmotion,li2018unsupervised} or use two-stream networks~\cite{jain2017fusionseg,tokmakov2017learning} to learn spatio-temporal representations in an end-to-end fashion. However, a major drawback of these approaches is that they fail to consider the importance of deep interactions between appearance and motion in learning rich spatio-temporal features. To address this issue, we propose a deep interleaved two-stream encoder, in which a motion transition module is leveraged for more effective representation learning.

\textbf{Neural Attention.}
Neural attention has been widely used in recent neural networks for various tasks, such as object recognition~\cite{hu2018squeeze,woo2018cbam,xie2019attentive}, re-identification~\cite{zhou2018aware}, visual saliency~\cite{wang2019salient} and medical imaging~\cite{zhou2019collaborative}. It allows the networks to focus on the most informative parts of the inputs. In this work, neural attention is used in two ways: first, in the encoder network, soft attention is applied independently to intermediate appearance or motion feature maps, and motion attention is further transferred to enhance the appearance attention. Second, in the bridge network, a scale-sensitive attention module is designed to obtain more compact features.


\begin{figure}[t]
	\centering
	\includegraphics[width=0.6\linewidth]{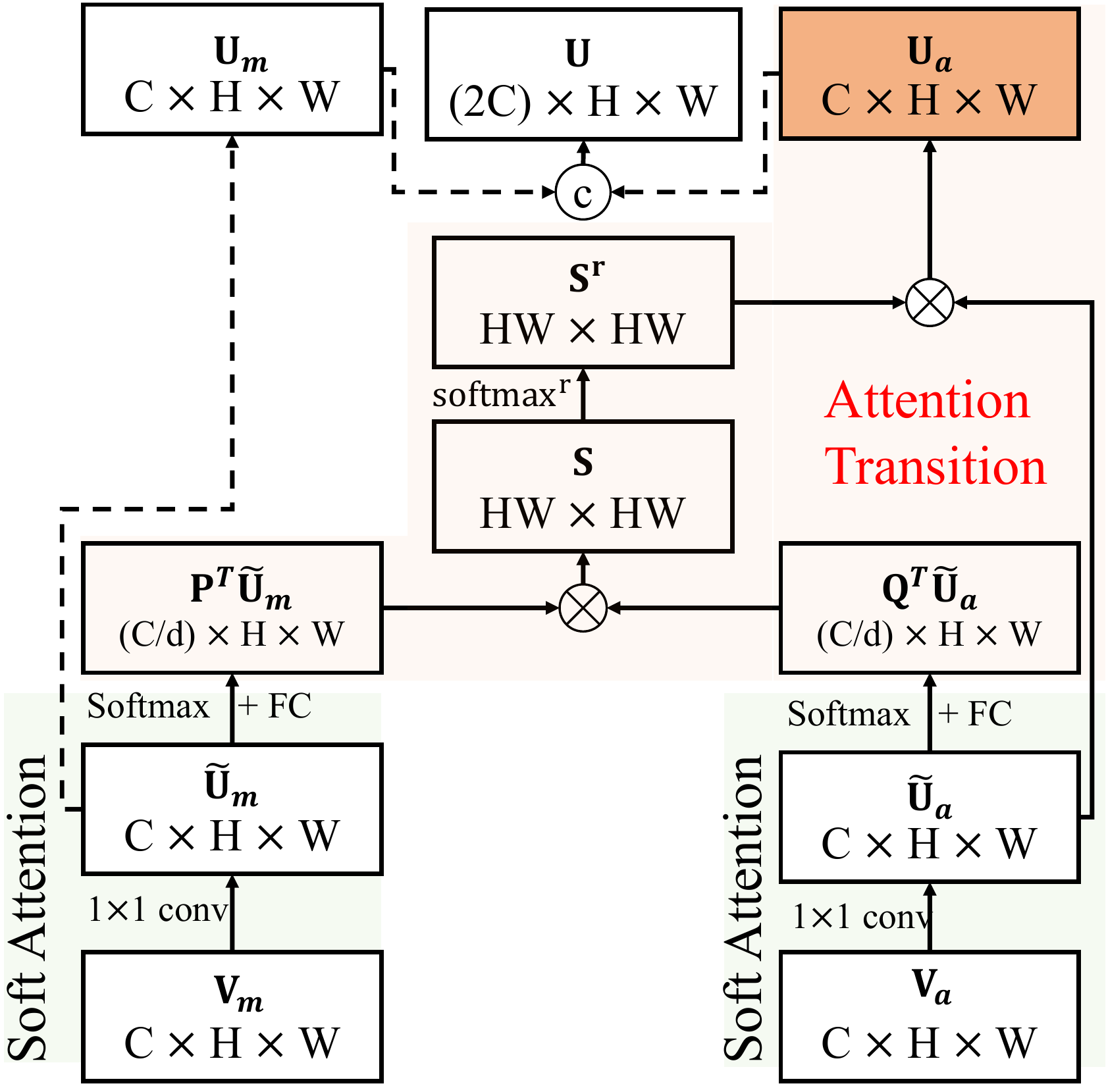}
	\caption{Computational graph of the MAT block. $\otimes$ and $\copyright$ indicate matrix multiplication and concatenation operations, respectively.}
	\vspace{-0.5cm}
	\label{fig:mat}
\end{figure}

\section{Proposed Method}

\subsection{Network Overview}

As illustrated in \figref{fig:flow}, MATNet is an end-to-end deep neural network for ZVOS, consisting of three concatenated networks, \ie, an interleaved encoder, a bridge network and a decoder.


\textbf{Interleaved Encoder Network.}
Our encoder relies on a two-stream structure to jointly encode object appearance and motion,  which has been proven effective in many related video tasks. Unlike previous works that treat the two streams equally, our encoder includes a MAT block at each network layer, which offers a motion-to-appearance pathway for information propagation.  To be specific, we take the first five convolutional blocks of ResNet-101~\cite{he2016deep} as the backbone for each stream. Given an RGB frame $\mathbf{I}_a$ and its optical flow map $\mathbf{I}_m$,
the encoder extracts intermediate features $\mathbf{V}_{a,i}\!\in\!\mathbb{R}^{W\times H \times C}$ and $\mathbf{V}_{m,i}\!\in\!\mathbb{R}^{W\times H \times C}$, respectively, at the $i$-th ($i\!\in\!\{2,3,4,5\}$) residual stage. The MAT block $\mathcal{F}_\text{MAT}$ enhances these features as follows:
\begin{equation}
\mathbf{U}_{a,i}, \mathbf{U}_{m,i} = \mathcal{F}_\text{MAT}(\mathbf{V}_{a,i}, \mathbf{V}_{m,i}),
\end{equation}
where $\mathbf{U}_{\cdot, i}\!\in\!\mathbb{R}^{W\times H \times C}$ indicates the enhanced features.
We then obtain the spatio-temporal object representation  $\mathbf{U}_i$ at the $i$-th stage as
$\mathbf{U}_i\!=\!\mathtt{Concat}(\mathbf{U}_{a,i},\!\mathbf{U}_{m,i})\!\in\! \mathbb{R}^{W\times H \times 2C}$ which is further fed into the down-stream decoder via a bridge network.


\textbf{Bridge Network.}
The bridge network is expected to selectively transfer encoder features to the decoder.
It is formed by SSA modules, each of which takes advantage of the encoder feature $\mathbf{U}_i$ at the $i$-th stage and predicts an attention-aware feature $\mathbf{Z}_i$ .
This is achieved by a two-level attention scheme, wherein
the local-level attention adopts  channel-wise and spatial-wise attention mechanisms to focus input features on the correct object regions
as well as suppress possible noises existing in the redundant features,
while the global-level attention aims to re-calibrate the features to account for objects of different sizes.


\textbf{Decoder Network.}
The decoder network takes a coarse-to-fine scheme to carry out segmentation.
It is formed by four  BAR modules, \ie, $\text{BAR}_i, i\!\in\!\{2,3,4,5\}$, each corresponding to the $i$-th residual block.
From $\text{BAR}_5$ to $\text{BAR}_2$, the resolution of feature maps gradually increases by compensating for high-level coarse features with more low-level details.
The $\text{BAR}_2$ produces the finest feature map, whose resolution is 1/4 of the input image size.
It is processed by two additional layers, $\mathtt{conv(3\!\times\!3,\!1)}\!\rightarrow\!\mathtt{sigmoid}$, to obtain the final mask output $\mathbf{M}^s\in\mathbb{R}^{W\times H}$.
%

%

As follows, we will introduce the three proposed modules (\ie, MAT, SSA, BAR) in detail.
For simplicity, we omit the subscript $i$.
%
%

%

\begin{figure}[t]
	\centering
	\includegraphics[width=\linewidth]{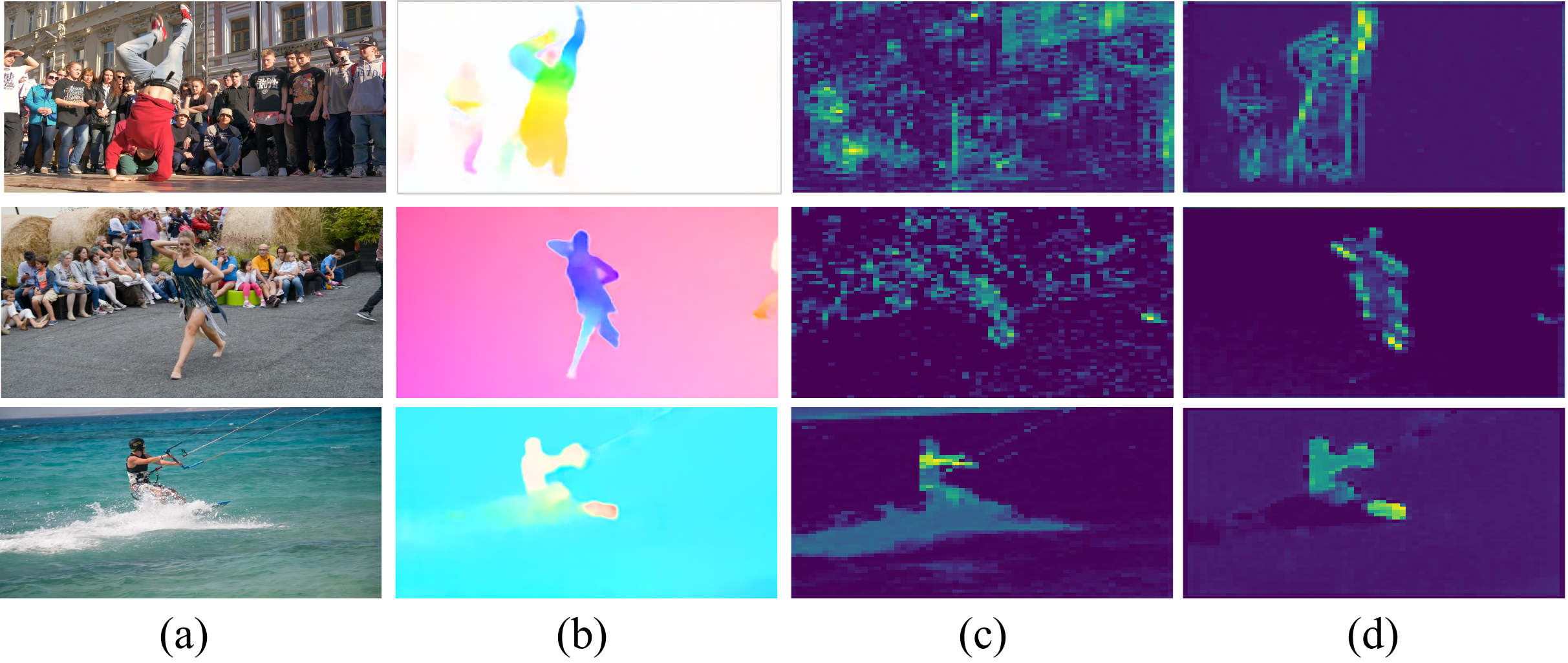}
	\caption{Illustration of the effects of our MAT block. (a) Image. (b) Optical flow. Comparing the  feature maps in $\mathbf{V}_a$ (c) and in $\mathbf{U}_a$ (d), we find that our MAT block can emphasize important regions and suppress background responses, providing more effective object representation for segmentation.}
	\label{fig:mat-vis}
\end{figure}

\subsection{Motion-Attentive Transition Module}

The MAT module is comprised of two units: a soft attention (SA) unit and an attention transition (AT) unit, as shown in \figref{fig:mat}.
The former unit helps to focus on the important regions of the inputs, while the latter transfers the attentive motion features to facilitate appearance learning.

\textbf{Soft Attention:}
This unit softly weights the input feature map $\mathbf{V}_m$  (or $ \mathbf{V}_a $) at each spatial location.
Taking $\mathbf{V}_m$ as the input,
the SA unit outputs a motion-attentive feature $\tilde{\mathbf{U}}_m \in \mathbb{R}^{W\times H \times C}$ as follows:
\begin{align}\label{eq:softmax}
\begin{split}
&\text{{\small Softmax attention:}} \    \mathbf{A}_m = \mathtt{softmax}(\mathbf{w}_m \ast \mathbf{V}_m), \\
&\text{{\small Attention-enhanced feature:}} \   \tilde{\mathbf{U}}_m^c = \mathbf{A}_m \Modot \mathbf{V}_m^c,
\end{split}
\end{align}
where $ \mathbf{w}_m\in\mathbb{R}^{1\times 1\times C} $ is a $1\!\times\!1$ conv kernel that maps $ \mathbf{V}_m $ to a significance matrix, which is normalized using $\mathtt{softmax}$ to achieve a soft attention map $\mathbf{A}_m\in\mathbb{R}^{W\times H}$.
$\ast$ indicates the conv operation.
$\tilde{\mathbf{U}}_m\in\mathbb{R}^{W\times H \times C}$ is the attention-aware feature map.
$\tilde{\mathbf{U}}_m^c$ and $ \mathbf{V}_m^c $ indicate the 2D feature slices of $\tilde{\mathbf{U}}_m$ and $ \mathbf{V}_m $ at the $c$-th channel, respectively.
$\Modot$ indicates the element-wise multiplication.
Similarly, given $\mathbf{V}_a$, we can obtain the appearance-attentive feature
$\tilde{\mathbf{U}}_a$ by \equationref{eq:softmax}.



\textbf{Attention Transition:}
To transfer motion-attentive features $\tilde{\mathbf{U}}_m$,  we first seek the affinity between $\tilde{\mathbf{U}}_a$ and $\tilde{\mathbf{U}}_m$ in a non-local manner using the following multi-modal bilinear model:
\begin{equation}\label{eq:1}
\mathbf{S} = \tilde{\mathbf{U}}_m^\top\mathbf{W}\tilde{\mathbf{U}}_a \in \mathbb{R}^{(WH)\times (WH)},
\end{equation}
where $\mathbf{W}\in \mathbb{R}^{C\times C}$ is a trainable weight matrix.
%
The affinity matrix $\mathbf{S}$ can effectively capture pairwise relationships between the two feature spaces. However, it also introduces a huge number of parameters, which increases the computational cost and creates the risk of over-fitting.  To overcome this problem, $\mathbf{W}$ is approximately factorized into two low-rank matrices $\mathbf{P}\in \mathbb{R}^{C\times\frac{C}{d}}$ and $\mathbf{Q}\in\mathbf{R}^{C\times\frac{C}{d}}$,  where $d$ $(d\!>\!1)$ is a reduction ratio. Then, \equationref{eq:1} can be rewritten as:
\begin{equation}\label{eq:2}
\mathbf{S} = \tilde{\mathbf{U}}_m^\top\mathbf{P}\mathbf{Q}^\top\tilde{\mathbf{U}}_a= (\mathbf{P}^\top\tilde{\mathbf{U}}_m)^\top(\mathbf{Q}^\top\tilde{\mathbf{U}}_a).
\end{equation}
This operation is equal to applying channel-wise feature transformations to $\tilde{\mathbf{U}}_m$ and $\tilde{\mathbf{U}}_a$
before computing the similarity.
This not only significantly reduces the number of parameters by 2/$d$ times,
but also generates a compact channel-wise feature representation for each modal.
Then, we normalize $\mathbf{S}$ row-wise to derive an attention map $\mathbf{S}^r$ conditioned on motion features
and achieve enhanced appearance features $\mathbf{U}_a\in\mathbb{R}^{W\times H\times C}$:
\begin{align}\label{eq:za}
\begin{split}
\text{\small Motion conditioned attention:} \quad & \mathbf{S}^r = \mathtt{softmax}^r(\mathbf{S}),  \\
\text{\small Attention-enhanced feature:} \quad & \mathbf{U}_a = \tilde{\mathbf{U}}_a\mathbf{S}^r,
\end{split}
\end{align}
where $\mathtt{softmax}^r$ indicates  row-wise $\mathtt{softmax}$.

\textit{Deep} MAT:
Deep network structures have achieved great success due to their powerful representational ability.
Therefore, we extend the MAT module into a deep structure consisting of $L$ MAT layers cascaded in depth (denoted by $\mathcal{F}_\text{MAT}^{(1)}$,$\mathcal{F}_\text{MAT}^{(2)}$, $\cdots$,$\mathcal{F}_\text{MAT}^{(L)}$).
Let $\mathbf{U}_a^{(l-1)}$ and $\mathbf{U}_m^{(l-1)}$ be the input features for $\mathcal{F}_\text{MAT}^{(l)}$.
It then outputs features $\mathbf{U}_a^{(l)}$ and $\mathbf{U}_m^{(l)}$, which are further fed to $\mathcal{F}_\text{MAT}^{(l+1)}$ in a recursive manner:
\begin{equation}\label{eq:stack}
\mathbf{U}_a^{(l)}, \mathbf{U}_m^{(l)} = \mathcal{F}_\text{MAT}^{(l)}(\mathbf{U}_a^{(l-1)}, \mathbf{U}_m^{(l-1)}),
\end{equation}
where $\mathbf{U}_a^{(l)}$ is computed as in \equationref{eq:za} and $\mathbf{U}_m^{(l)}\!=\!\tilde{\mathbf{U}}_m^{(l-1)}$ following \equationref{eq:softmax}. In addition, we have $\mathbf{U}_a^{(0)}\!=\!\mathbf{V}_a$ and $\mathbf{U}_m^{(0)}\!=\!\mathbf{V}_m$.

It is worth noting that stacking MAT modules directly leads to an obvious performance drop.
Inspired by~\cite{wang2017residual},
we propose stacking multiple MAT modules in a residual form by modifying the outputs of \equationref{eq:stack} as follows:
\begin{align}
\begin{split}
\mathbf{U}_a^{(l)} &= \mathbf{U}_a^{(l-1)} + \tilde{\mathbf{U}}_a^{(l-1)}\mathbf{S}^r \\
&= \mathbf{U}_a^{(l-1)} + (\mathbf{A}_a^{(l-1)}\Modot\mathbf{V}_a^{(l-1)})\mathbf{S}^r.\\
\mathbf{U}_m^{(l)} &= \mathbf{U}_m^{(l-1)} + \tilde{\mathbf{U}}_m^{(l-1)} \\
&= \mathbf{U}_m^{(l-1)} + \mathbf{A}_m^{(l-1)}\Modot\mathbf{V}_m^{(l-1)}.
\end{split}	
\end{align}
Here, we combine \equationref{eq:softmax} and \equationref{eq:za} to provide a global view of our deep residual MAT modules.

In \figref{fig:mat-vis}, we show the effects of our MAT block. We see that the features in $\mathbf{V}_a$ (\figref{fig:mat-vis} (c)) are well refined by the MAT block to produce features in $\mathbf{U}_a$ (\figref{fig:mat-vis} (d)).

\begin{figure}[t]
	\centering
	\includegraphics[width=\linewidth]{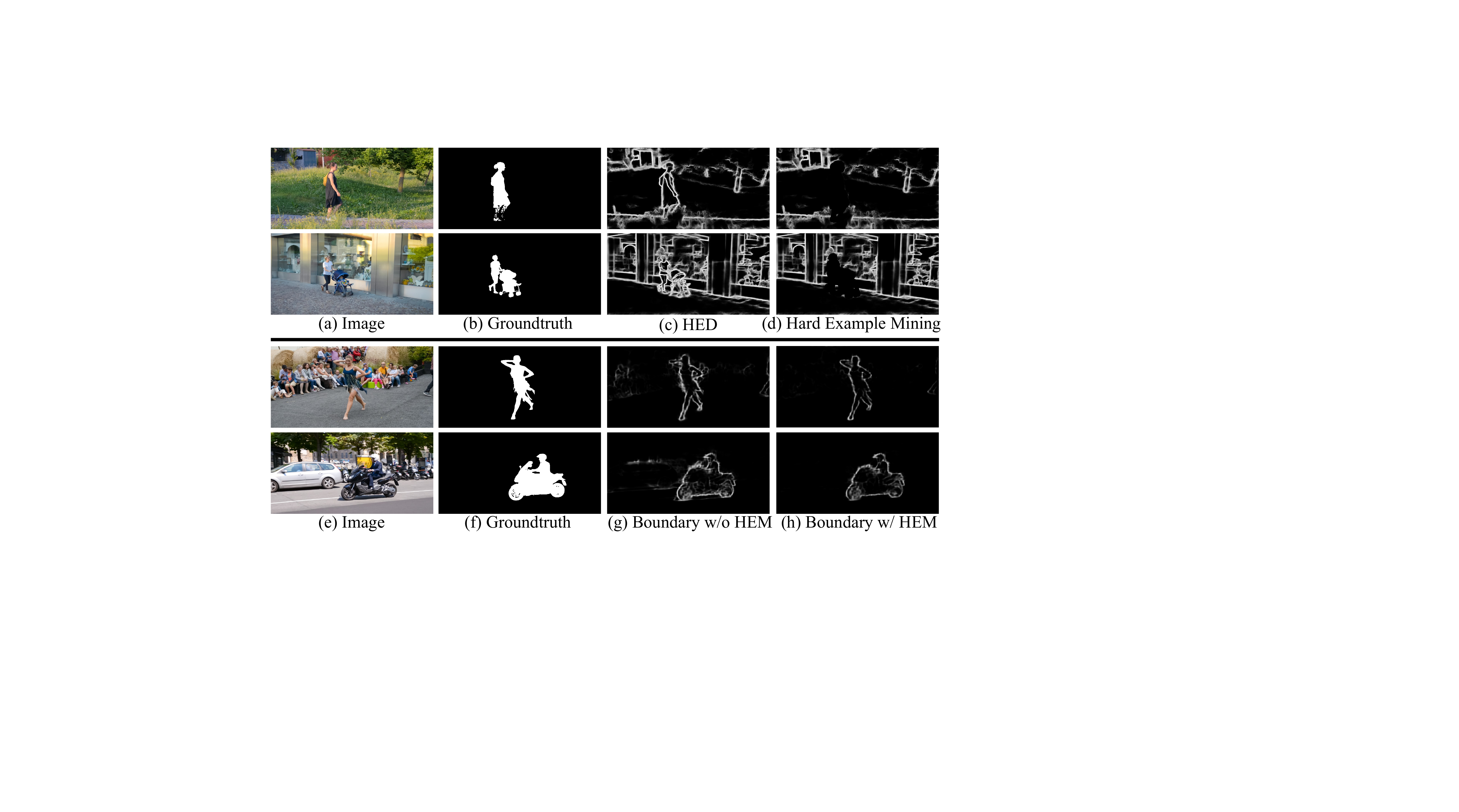}
	\caption{Illustration of hard example mining (HEM) for object boundary detection. During training, for each training image in (a), our method first estimates an edge map (c) using off-the-shelf HED~\cite{xie2015holistically}, and then determines hard pixels (d) to facilitate training. For each test image in (e), we see that the boundary results with HEM (h) are more accurate than those without HEM (g).}
	\label{fig:edge}
\end{figure}
\subsection{Scale-Sensitive Attention Module}

%

The SSA module is extended from a simplified CBAM $\mathcal{F}_\text{CBAM}$~\cite{woo2018cbam} by adding a global-level attention $\mathcal{F}_g$.
Given a feature map $\mathbf{U}\in\mathbb{R}^{W\times H \times 2C}$, our SSA module refines it
as follows:
\begin{equation}
\mathbf{Z} = \mathcal{F}_g(\mathcal{F}_{\text{CBAM}}(\mathbf{U}))\in\mathbb{R}^{W\times H \times 2C}.
\end{equation}

The CBAM module $\mathcal{F}_\text{CBAM}$ consists of two sequential sub-modules: channel and spatial attention, which can be formulated as:
\begin{align}\label{eq:cbam}
\begin{split}
\text{\small Channel attention:}~~&\mathbf{s} = \mathcal{F}_s(\mathbf{\mathbf{U}}),
~~ \mathbf{e} = \mathcal{F}_e(\mathbf{s}), \\
&\mathbf{Z}_c=\mathbf{e} \star \mathbf{U}, \\
\text{\small Spatial attention:}~~ & \mathbf{p} = \mathcal{F}_p(\mathbf{Z}_c),~\mathbf{Z}_\text{CBAM} = \mathbf{p} \Modot \mathbf{Z}_c,
\end{split}
\end{align}
where $\mathcal{F}_s$ is a \textit{squeeze} operator that gathers the global spatial information of $\mathbf{U}$ into a vector $\mathbf{s}\in\mathbb{R}^{2C}$, while
$\mathcal{F}_e$ is an \textit{excitation} operator that captures channel-wise dependencies and outputs an attention vector $\mathbf{e}\in\mathbb{R}^{2C}$.
Following~\cite{hu2018squeeze}, $\mathcal{F}_s$ is implemented by applying $ \mathtt{avg pooling} $ on each feature channel, and $\mathcal{F}_e$ is formed by four consecutive operations: $\mathtt{fc(\frac{2C}{16}) \rightarrow ReLU \rightarrow fc(2C)  \rightarrow sigmoid}$.
$\mathbf{Z}_c\in\mathbf{R}^{W\times H \times 2C}$ denotes channel-wise attentive features, and $\star$ indicates the channel-wise multiplication.
In the spatial attention, $\mathcal{F}_p $ exploits the inter-spatial relationship of $\mathbf{Z}_c$ and produces a spatial-wise attention map $\mathbf{p}\in\mathbb{R}^{W\times H}$ by $\mathtt{conv(7\times7,1) \rightarrow sigmoid}$.
Then, we achieve the attention glimpse $\mathbf{Z}_\text{CBAM}\in\mathbb{R}^{W\times H \times 2C}$ as the local-level feature.

%

The global-level attention $\mathcal{F}_g$ shares a similar spirit to the channel attention layer in \equationref{eq:cbam},
in that it shares the same \textit{squeeze} layer but modifies the \textit{excitation} layer as $\mathtt{fc(\frac{2C}{16}) \rightarrow fc(1) \rightarrow sigmoid}$ to output a scale-selection factor $\mathbf{g}\in\mathbb{R}^1$ and then obtain scale-sensitive features $\mathbf{Z}$ as follows:
\begin{equation}\label{eq:im}
\mathbf{Z} = (\mathbf{g}*\mathbf{Z}_\text{CBAM}) + \mathbf{U}.
\end{equation}
Note that we use identity mapping to avoid losing important information on the regions with attention values close to 0.

\subsection{Boundary-Aware Refinement Module}

\begin{figure}[t]
	\centering
	\includegraphics[width=0.7\linewidth]{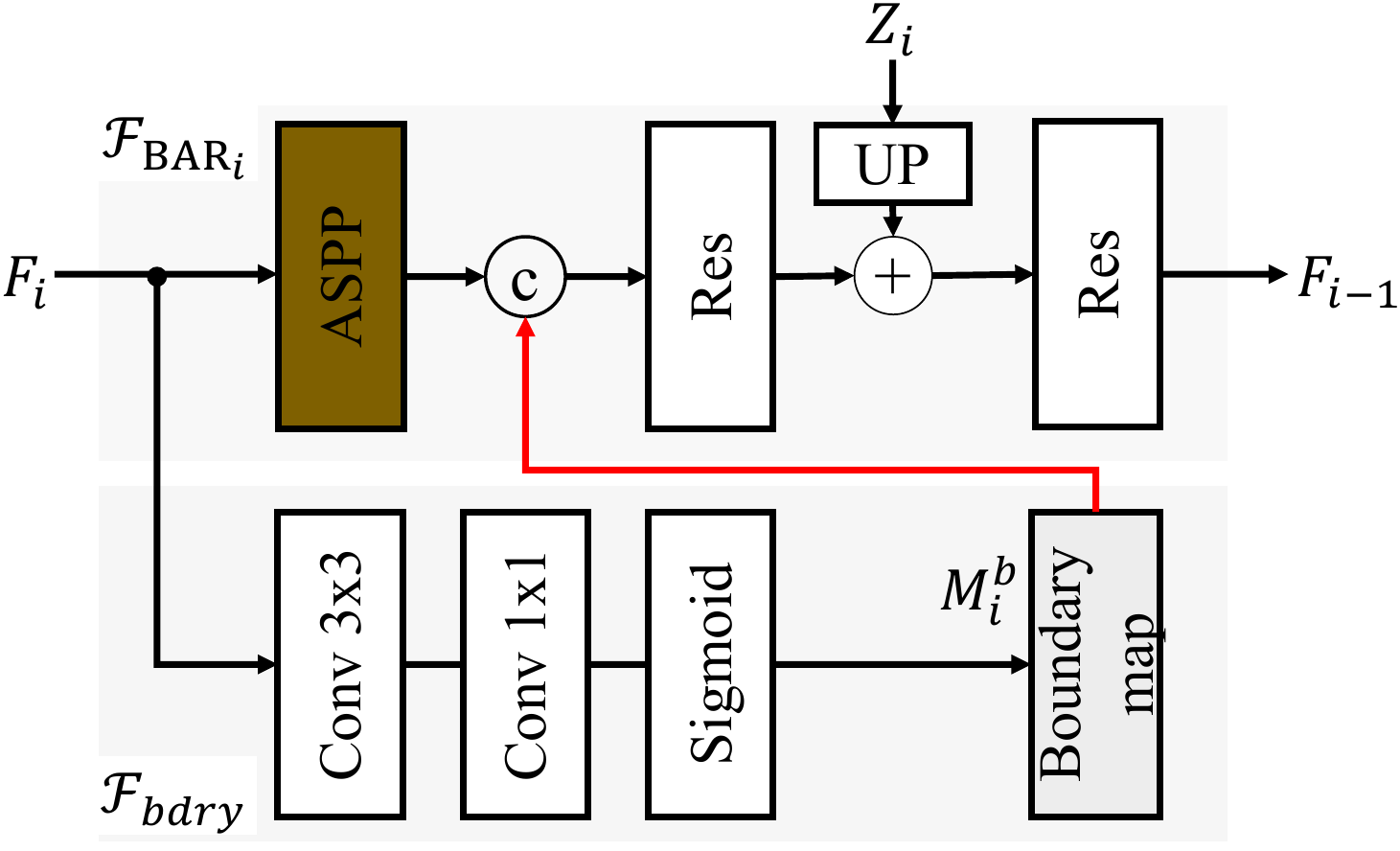}
	\caption{Computational graph of the $\text{BAR}_i$ block. $\copyright$ and $\oplus$ indicate concatenation and element-wise addition operations, respectively.
	}
	\label{fig:bar}
\end{figure}

In the decoder network, each BAR module, \eg, $\text{BAR}_i$, receives two inputs, \ie, $\mathbf{Z}_{i}$ from the corresponding SSA module and $\mathbf{F}_{i}$ from the previous BAR.
To obtain a sharp mask output, the BAR first performs object boundary estimation using an extra boundary detection module $\mathcal{F}_\text{bdry}$, which compels the network to emphasize finer object details.
The predicted boundary map is then combined with the two inputs to produce finer features for the next BAR module.
It can be formulated as:
\begin{align}
\begin{split}
\mathbf{M}_i^b &= \mathcal{F}_\text{bdry} (\mathbf{F}_{i}), \\
\mathbf{F}_{i-1} &= \mathcal{F}_{\text{BAR}_i}(\mathbf{Z}_{i}, \mathbf{F}_{i}, \mathbf{M}_i^b),
\end{split}
\end{align}
where $\mathcal{F}_\text{bdry}$ consists of a stack of convolutional layers and a sigmoid layer,  $\mathbf{M}_i^b\in\mathbb{R}^{W\times H}$ indicates the boundary map and
$\mathbf{F}_{i-1}$ is the output feature map of $\text{BAR}_i$. The computational graph of $\text{BAR}_i$ is shown in \figref{fig:bar}.

BAR benefits from two key factors:
the first is that we apply \textit{Atrous Spatial Pyramid Pooling} (ASPP)~\cite{chen2017deeplab} on convolutional features to transform them into a multi-scale representation. This helps to enlarge the receptive field and obtain more spatial details for decoding.

The second benefit is that we introduce a heuristic method for automatically mining hard negative pixels to support the training of $\mathcal{F}_\text{bdry}$.
Specifically, for each training frame,
we use the popular off-the-shelf HED model~\cite{xie2015holistically} to predict a boundary map $\mathbf{E}\in[0, 1]^{W\times H}$, wherein each value $\mathbf{E}_k$ represents the probability of pixel $k$ being an edge pixel.
Then, pixel $k$ is regarded as a hard negative pixel if it has a  high edge probability (\eg, $\mathbf{E}_k\!>\!0.2$) and falls outside the dilated ground-truth region. If pixel $k$ is a hard pixel, then its weights $w_k\!=\!1 + \mathbf{E}_k$; otherwise, $w_k\!=\!1$.
%
%

Then, $w_k$ is used to weight the following boundary loss so that it can be penalized heavily if the hard pixels are misclassified:
\begin{align}\label{loss}
\begin{split}
\!\!\!\!\mathcal{L}_\text{bdry}(\mathbf{M}^b, \mathbf{G}^b) \!=\! -\!\sum\nolimits_k w_k (&(1\!-\!\mathbf{G}_k^b)\log(1\!-\!\mathbf{M}_k^b)\\
\!&+ \mathbf{G}_k^b\log(\mathbf{M}_k^b)),\!\!\!
\end{split}
\end{align}
where $\mathbf{M}^b$ and $\mathbf{G}^b$ are the boundary prediction and  ground-truth, respectively.

\figref{fig:edge} offers an illustration of the above hard example mining (HEM) scheme.
Clearly,  by explicitly discovering hard negative pixels, the network can produce more accurate boundary prediction with well-suppressed background pixels (see \figref{fig:edge} (g) and (h)).

\begin{figure*}
	\centering
	\includegraphics[width=\linewidth]{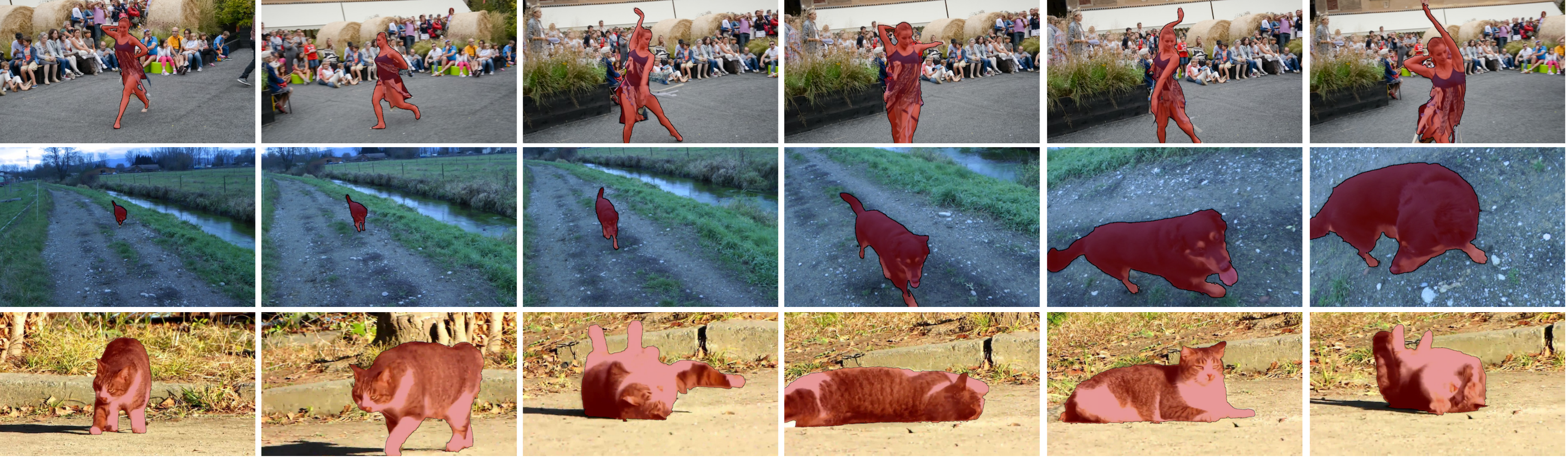}
	\caption{Qualitative results on three sequences. From top to bottom: \textit{dance-twirl} from DAVIS-16, \textit{dogs02} from  FBMS, and \textit{cat-0001} from Youtube-Objects.}
	\vspace{-0.5cm}
	\label{fig:image}
\end{figure*}

\subsection{Implementation Details}

\textbf{Training Loss.}
Given an input frame $\mathbf{I}_a\in\mathbb{R}^{473\times 473\times 3}$,
our MATNet predicts a segmentation mask $\mathbf{M}^s\in[0,1]^{473\times 473}$ and four boundary predictions $\{\mathbf{M}_i^b \in [0,1]^{473\times 473}\}_{i=1}^4$ using BAR modules.
Let $\mathbf{G}^s\!\in\!\{0,1\}^{473\!\times\!473}$ be the binary segmentation ground-truth, and $\mathbf{G}^b\!\in\!\{0,1\}^{473\!\times\!473}$ be the boundary ground-truth which can be easily computed from $\mathbf{G}^s$.
The overall loss function is formulated as:
\begin{equation}
\mathcal{L} = \mathcal{L}_\text{CE}(\mathbf{M}^s, \mathbf{G}^s) + \frac{1}{N}\sum\nolimits_{i=1}^{N=4} \mathcal{L}_\text{bdry}(\mathbf{M}_i^b, \mathbf{G}^b),
\end{equation}
where $ \mathcal{L}_\text{CE} $ indicates the classic cross entropy loss.

\noindent\textbf{Training Settings.}
We train the proposed neural network in an end-to-end manner.
Our training data consist of two parts:
i) all training data in DAVIS-16~\cite{perazzi2016benchmark}, which includes 30 videos with about 2K frames;
ii) a subset of 12K frames selected from the training set of Youtube-VOS~\cite{xu2018youtube}, which is obtained by sampling images every ten frames in each video.
In total, we have 14K training samples, basically matching AGS~\cite{Wang_2019_CVPR}, which uses 13K training samples.
For each training image of size $473\!\times\!473\!\times 3$,
we first estimate its optical flow using PWC-Net~\cite{sun2018pwc} due to its high efficiency and accuracy.
The entire network is trained using the SGD optimizer with an initial learning rate of 1e-4 for the encoder and the bridge network, and 1e-3 for the decoder.
During training, the batch size, momentum and weight decay are set to $2$, $0.9$, and 1e-5, respectively.
The data are augmented online with horizontal flip and rotations covering a range of degrees $(-10,10)$.
The network is implemented with PyTorch,
and all the experiments are conducted using a single Nvidia RTX 2080Ti GPU and an Intel(R) Xeon Gold 5120 CPU.

\noindent\textbf{Test Settings.}
Once the network is trained, we apply it to unseen videos.
Given a test video, we resize all the frames to $473\!\times\!473$, and feed each frame, along with its optical flow, to the network for segmentation.
We follow the common protocol used in previous works~\cite{Wang_2019_CVPR,perazzi2017learning,xiao2018monet} and employ  CRF to obtain the final binary segmentation results.

\noindent\textbf{Runtime.}
For each test image of size $473\!\times\!473\!\times 3$, the forward inference of our MATNet takes about 0.05s,
while optical flow estimation and CRF-based post-processing take about 0.2s and 0.5s, respectively.

\section{Experiments}

\subsection{Experimental Setup}
We carry out comprehensive experiments on three popular datasets: DAVIS-16~\cite{perazzi2016benchmark}, Youtube-Objects~\cite{prest2012learning} and FBMS~\cite{ochs2013segmentation}.

\textbf{DAVIS-16} consists of 50 high-quality video sequences (30 for training and 20 for validation) in total.
Each frame contains pixel-accurate annotations for foreground objects.
For quantitative evaluation, we use three standard metrics suggested by~\cite{perazzi2016benchmark}, namely region similarity $\mathcal{J}$, boundary accuracy $\mathcal{F}$, and time stability $\mathcal{T}$.

\textbf{Youtube-Objects} is a large dataset of 126 web videos with 10 semantic object categories and more than 20,000 frames. Following its protocol,
we use the region similarity $\mathcal{J}$ metric to measure the performance.

\textbf{FBMS} is composed of 59 video sequences with ground-truth annotations provided in a subset of the frames.
Following the standard protocol~\cite{tokmakov2017learning}, we do not use any sequences for training and only evaluate on the validation set consisting of 30 sequences.
%

\subsection{Ablation Study}

\tabref{table:ablation} summarizes the ablation analysis of our MATNet on DAVIS-16.
%

\textbf{MAT Block.}
We first study the effects of the MAT block by comparing our full model to one of the same architecture without MAT, denoted as MATNet \textit{w/o} MAT.
The encoder in this network is thus equivalent to a standard two-stream model, where convolutional features from the two streams are concatenated at each residual stage for object representation.
As shown in \tabref{table:ablation}, this model encounters a huge performance degradation ($2.9\%$ in Mean $\mathcal{J}$ and $3.4\%$ in Mean $\mathcal{F}$), which demonstrates the effectiveness of the MAT block.

Moreover, we also evaluate the performance of MATNet with a different number of MAT blocks in each deep residual MAT layer.
As shown in \tabref{table:L}, the performance of the model gradually improves as $L$ increases, reaching saturation at $L=5$.
Based on this analysis, we use $L=5$ as the default number of MAT blocks in MATNet.

\textbf{SSA Block.}
To measure the effectiveness of the SSA block, we design another network variant, MATNet \textit{w/o} SSA, by replacing the SSA block with a simple skip layer.
As can be observed, its performance is $1.7\%$ lower than our full model in terms of Mean $\mathcal{J}$, and $1.0\%$ lower in Mean $\mathcal{F}$.
The performance drop is mainly caused by the redundant spatio-temporal features from the encoder.
Our SSA block aims to eliminate the redundancy by only focusing on the features that are beneficial to segmentation.

\begin{table}[t]
	\centering
	\small
	\caption{Ablation study of the proposed network on DAVIS-16, measured by the Mean $\mathcal{J}$ and Mean $\mathcal{F}$.}
	\begin{tabular}{l||cc|cc}
		\hlineB{2.5}
		Network Variant & Mean $\mathcal{J}\uparrow$ & $\Delta \mathcal{J}$ & Mean $\mathcal{F}\uparrow$ & $\Delta \mathcal{F}$ \\
		\hline \hline
		MATNet \textit{w/o} MAT     & 79.5 & -2.9 & 77.3 & -3.4 \\
		MATNet \textit{w/o} SSA     & 80.7 & -1.7 & 79.7 & -1.0 \\
		MATNet \textit{w/o} HEM     & 81.4 & -1.0 & 78.4 & -2.3     \\
		MATNet \textit{w/ } Res50   & 81.1 & -1.3 & 79.3 & -1.4 \\ \hline
		MATNet \textit{w/ } Res101  & \textbf{82.4} & -    & \textbf{80.7} & - \\
		\hlineB{2.5}
	\end{tabular}
	
	\label{table:ablation}
\end{table}

\begin{table}[t]
	\centering
	\small
	\caption{Performance comparisons with different numbers of MAT blocks cascaded in each MAT layer on DAVIS-16. }
	\begin{tabular}{c||ccccc}
		\hlineB{2.5}
		Metric		&$L=0$ & $L=1$ & $L=3$ & $L=5$ & $L=7$ \\ \hline \hline
		Mean $\mathcal{J}\uparrow$	&   79.5  &  80.6     &   81.6    &    82.4   &   82.2    \\ \hline
		Mean $\mathcal{F}\uparrow$	&   77.3  &  80.3     &   80.7    &    80.7   &   80.6   \\
		\hlineB{2.5}
	\end{tabular}
	
	\label{table:L}
\end{table}

\begin{table*}[t]
	\small
	\centering
	\caption{Quantitative comparison of ZVOS methods on the DAVIS-16 validation set.
		The best result for each metric is \textbf{boldfaced}.
		All the results are borrowed from the public leaderboard maintained by the DAVIS challenge.}
	\begin{tabular}{cr||cccccccccccc}
		\hlineB{2.5}
		& Measure                                    & SFL & FSEG &LVO   &ARP   &PDB   &LSMO  &MotAdapt  & EPO &AGS  & COSNet  & AGNN & \textbf{MATNet}          \\ \hline \hline
		& Mean$\uparrow$                       &67.4 & 70.7  &75.9  &76.2  &77.2  &78.2  &77.2  & 80.6  &79.7  & 80.5 & 80.7 & \textbf{82.4}   \\
		$\mathcal{J}$	& Recall$\uparrow$  & 81.4 & 83.5  &89.1  &91.1  &90.1  &89.1  &87.8   & \textbf{95.2}   &91.1  & 93.1 & 94.0&  94.5  \\
		& Decay$\downarrow$                 &6.2 & 1.5   &\textbf{0.0}   &7.0   &0.9   &4.1   &5.0   & 2.2  &1.9  & 4.4 &\textbf{0.0}  & 5.5            \\
		\hline
		& Mean$\uparrow$                        &66.7 & 65.3  &72.1  &70.6  &74.5  &75.9  &77.4  & 75.5  &77.4  &79.5  & 79.1& \textbf{80.7}  \\
		$\mathcal{F}$ 	& Recall$\uparrow$   &77.1 & 73.8  &83.4  &83.5  &84.4  &84.7  &84.4 & 87.9    &85.8 & 89.5  & \textbf{90.5} & 90.2  \\
		& Decay$\downarrow$                   &5.1 & 1.8   &1.3   &7.9   &\textbf{-0.2}  &3.5   &3.3   &2.4    &1.6 & 5.0  & 0.0 & 4.5            \\ \hline
		$\mathcal{T}$	& Mean$\downarrow$  &28.2 & 32.8  &26.5  &39.3  &29.1  &21.2  &27.9   &19.3   &26.7 & \textbf{18.4} & 33.7  & 21.6           \\
		\hlineB{2.5}
	\end{tabular}
	
	\label{table:davis16}
\end{table*}

\textbf{Effectiveness of HEM.}
We also study the influence of using HEM during training.
HEM is expected to facilitate the learning of more accurate object boundaries, which should further boost the segmentation procedure.
The results in \tabref{table:ablation} (see MATNet \textit{w/o} HEM) indicate the importance of HEM.
By directly controlling the loss function in \equationref{loss}, HEM helps to improve the contour accuracy by $2.3\%$.

\textbf{Impact of Backbone.} To verify that the high performance of our network is not mainly due to the powerful backbone, we replace ResNet-101 with ResNet-50 to construct another network, \ie, MATNet \textit{w/} Res50.
We see that the performance slightly degrades, but it still outperforms AGS in terms of both Mean $\mathcal{J}$ and Mean $\mathcal{F}$.
This further confirms the effectiveness of MATNet.

\begin{figure}[t]
	\centering
	\includegraphics[width=\linewidth]{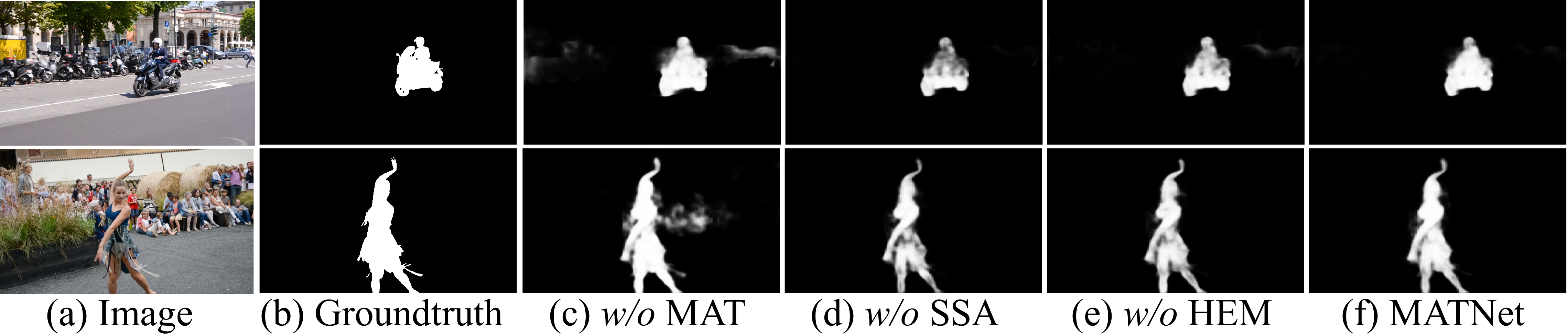}
	\vspace{-10pt}
	\caption{Qualitative results of ablation study.}
	\vspace{-5pt}
	\label{fig:ablation}
\end{figure}

\textbf{Qualitative Comparison.}
\figref{fig:ablation} shows visual results of the above ablation studies on two sequences. We see that all of the network variants produce worse results compared with MATNet. It should also be noted that the MAT block has the greatest impact on the performance.

%
%
\begin{table}[t]
	
	\setlength{\tabcolsep}{0.4em}	
	\centering
	\small
	\caption{Quantitative results for each category on Youtube-Objects over Mean $\mathcal{J}$. }
	
	\begin{tabular}{c||cccccccc}
		\hlineB{2.5}
		Category  &  LVO & SFL &FSEG & PDB  &AGS   & \textbf{MATNet}\\ \hline \hline
		Airplane  &86.2 &65.6 &81.7 &78.0  &87.7  & 72.9\\
		Bird     &81.0 &65.4 &63.8 &80.0  &76.7  & 77.5\\
		Boat      &68.5 &59.9 &72.3 &58.9  &72.2  & 66.9\\	
		Car       &69.3 &64.0 &74.9 &76.5  &78.6  & 79.0\\
		Cat       &58.8 &58.9 &68.4 &63.0  &69.2  & 73.7\\
		Cow       &68.5 &51.2 &68.0 &64.1  &64.6  & 67.4\\
		Dog       &61.7 &54.1 &69.4 &70.1  &73.3  & 75.9\\
		Horse     &53.9 &64.8 &60.4 &67.6  &64.4  & 63.2\\
		Motorbike &60.8 &52.6 &62.7 &58.4  &62.1  & 62.6\\
		Train     &66.3 &34.0 &62.2 &35.3  &48.2  & 51.0\\
		\hline
		Mean $\mathcal{J}\uparrow$      &67.5 &57.1 &68.4 &65.5  &\textbf{69.7}  & 69.0\\
		\hlineB{2.5}
	\end{tabular}
	
	\label{table:youtube}
\end{table}

\subsection{Comparison with State-of-the-arts}
\textbf{Evaluation on DAVIS-16.}
We compare our MATNet with the top performing ZVOS methods in the public leaderboard of DAVIS-16.
The detailed results are shown in  \tabref{table:davis16}.
We see that MATNet outperforms all the reported methods across most metrics.
Compared with the second-best, AGNN~\cite{wang2019zero}, MATNet obtains improvements of \textbf{1.7\%} and \textbf{1.6\%} in terms of Mean $\mathcal{J}$   and Mean $\mathcal{F}$, respectively.
%
%
In \tabref{table:davis16}, some of the deep learning-based models, \eg, FSEG~\cite{jain2017fusionseg}, LVO~\cite{tokmakov2017learning}, MoTAdapt~\cite{siam2019video} and EPO~\cite{faisal2019exploiting}, use motion cues to improve segmentation.
Our MATNet outperforms all of these methods by a large margin.
The reason lies in that  these methods learn motion and appearance features independently, without considering the close interactions between them.
In contrast, our MATNet can learn more effective multi-modal object representation with the interleaved encoder.
%

%

\textbf{Evaluation on Youtube-Objects.} \tabref{table:youtube} reports the detailed results on Youtube-Objects.
Our model also shows favorable performance, second only to AGS.
The performance gap is mainly caused by sequences in the Airplane and Boat categories, which contain objects that move very slowly and have visually similar appearances to their surroundings.
Both factors result in inaccurate estimation of optical flow.
In other categories, our model obtains a consistent performance improvement in comparison with AGS.


\begin{table}[t]
	\small
	\centering
	\setlength{\tabcolsep}{0.3em}	
	\caption{Quantitative results on FBMS over Mean $\mathcal{J}$.}
	\begin{tabular}{c||cccccccc}
		\hlineB{2.5}
		Measure		& ARP & MSTP	& FSEG & IET& OBN& PDB & \textbf{MATNet}\\ \hline
		Mean $\mathcal{J}\uparrow$ 	& 59.8 & 60.8 & 68.4& 71.9& 73.9& 74.0& \textbf{76.1}\\
		\hlineB{2.5}
	\end{tabular}
	
	\label{table:fbms}	
\end{table}

\textbf{Evaluation on FBMS.}
For completeness, we also evaluate our method on FBMS.
As shown in \tabref{table:fbms}, MATNet produces the best results with $\textbf{76.1\%}$ over Mean $\mathcal{J}$, which outperforms the second-best result, \ie, PDB~\cite{song2018pyramid}, by $\textbf{2.1\%}$.

\textbf{Qualitative Results.}
\figref{fig:image} depicts sample results for representative sequences from the three datasets.
The \textit{dance-twirl} sequence from DAVIS-16 contains many challenging factors, such as object deformation, motion blur and background clutter.
We can see that our method is robust to these challenges and delineates the target with accurate contours.
The effectiveness is further proved in \textit{cat-0001} from Youtube-Objects, in which the cat has a similar appearance to the surroundings and encounters large deformation.
In addition, our model also works well in \textit{dogs02}, in which the target suffers from large scale variations.

\section{Conclusion}
Inspired by the inherent multi-modal perception mechanism in HVS,
we present a novel model, MATNet, for ZVOS,
which introduces a new way of learning rich spatio-temporal object features.
This is achieved by MAT blocks within a two-stream interleaved encoder, which allow the transition of attentive motion features to enhance appearance learning at each convolution stage.
The encoder features are further processed by a bridge network to produce a compact and scale-sensitive representation, which is fed into a decoder to obtain accurate segmentation in a top-down fashion.
Extensive experimental results indicate that MATNet achieves favorable performance against current state-of-the-art methods.
The proposed interleaved encoder is a novel two-stream framework for spatio-temporal representation learning in videos,
and can be easily extended to other video analysis tasks, such as action recognition and video classification.

\bibliographystyle{aaai}
\small\bibliography{reference}

\end{document}